\documentclass[]{spie}  

\usepackage{amsmath,amsfonts,amssymb}
\usepackage{graphicx}
\usepackage{float}
\usepackage[colorlinks=true, allcolors=blue]{hyperref}
\usepackage[ruled,linesnumbered]{algorithm2e}

\title{Reproducing Kernel Hilbert Space Pruning for Sparse Hyperspectral Abundance Prediction}

\author[1]{Michael G. Rawson}
\author[1]{Timothy Doster}
\affil[1]{\{first\}.\{last\}@pnnl.gov,
Pacific Northwest National Laboratory, Seattle, WA, USA}

\begin{document} 
\maketitle

\begin{abstract}
Hyperspectral measurements from long range sensors can give a detailed picture of the items, materials, and chemicals in a scene but analysis can be difficult, slow, and expensive due to high spatial and spectral resolutions of state-of-the-art sensors. As such, sparsity is important to enable the future of spectral compression and analytics. It has been observed that environmental and atmospheric effects, including scattering, can produce nonlinear effects posing challenges for existing source separation and compression methods. 
We present a novel transformation into Hilbert spaces for pruning and constructing sparse representations via non-negative least squares minimization. Then we introduce max likelihood compression vectors to decrease information loss. Our approach is benchmarked against standard pruning and least squares as well as deep learning methods. Our methods are evaluated in terms of overall spectral reconstruction error and compression rate using real and synthetic data. We find that pruning least squares methods converge quickly unlike matching pursuit methods. We find that Hilbert space pruning can reduce error by as much as 40\% of the error of standard pruning and also outperform neural network autoencoders. 
\end{abstract}


\section{INTRODUCTION}

\quad Hyperspectral imaging has been used for decades to analyze materials, gases, and more. A spectrometer or hyperspectral imager measures the reflectivity/emmisivity of objects in a scene.   A measurement or signal collected from a scene can then be decomposed into a sum of labeled signature spectra, each referencing a known chemical compound, with varying precision. Once calculated, we call the coefficients of the signature spectra the `abundances' of that compound. This gives the chemistry of objects in scene which aids in classification and prediction of potential dangers. Now, finding the coefficients in a sum is a linear problem that can be solved with linear algebra. In order to have a unique solution, the signature spectra must be linearly independent. If the linear problem is underdetermined, that is does not have a unique solution, then we regularize to find the sparsest solution. 

Electromagnetic (EM) waves sum which gives us a linear problem. However, this is ignoring scattering effects. Scatter happens as EM waves reflect from material to material and eventually reach the hyperspectral sensor\cite{Unmixing}. From the partial differential equation of radiative transfer, scattering is a pointwise product. Given a spectral resolution, we can approximate the infinite pointwise product with the Hadamard, or elementwise, product of vectors. Every additional scattering event is an additional product with a signature spectrum. Since Hadamard products are commutative, if we have the signature spectra then we can calculate the scattering signature spectra that are produced by taking every possible product. 
Then we combine the signature spectra and product signature spectra in a weighted sum to represent the measurement. So the nonlinear problem has been turned into a linear problem that is solved with linear algebra. There are other nonlinear effects to account for however such as diffraction. One approximation is to use Wasserstein distances\cite{emerson2019path} to find close solutions with a nonstandard metric. 

In normal settings, it is safe to assume that the number of unique compounds present in a sample or pixel should be few. This sparsity assumption on abundances can be used to find unique solutions as we mentioned above but can also increase robustness to noise and decrease computational cost. There are many pruning methods to do this. The standard pruning method calculates the correlations between the signal and the signature spectra which follows from basis expansion. Then a threshold is set and low correlation signature spectra are discarded. Now, the correlation is the normalized inner product. However, we propose generalizing to consider other kernels. By mapping these signals to more general Hilbert spaces, the pruning becomes more robust to noise or errors, as we will see. We choose a Gaussian radial basis function (RBF) for a kernel accompanying a Hilbert space. This kernel value is then used for the thresholding in pruning, see all details in Section \ref{sec:meth}. 

Hyperspectral imaging can resolve hundreds of frequencies over a large spatial area resulting in gigabytes of recorded information; this problem is exacerbated by considering a sequence of images taken over time.  There has been significant work in the area of hyperspectral compression {\cite{Yang,Babu,ZHANG2015358}} to address this problem. Calculating sparse abundance vectors is a compression of the measurements at each pixel which also creates an interpretable product of scene constituents. It is possible that the abundance vector does not represent the measurement if the spectrum signature is not in the dictionary. Even if the spectrum signature is identified later, the data is lost and the spectrum signature or material class cannot be confirmed. Keeping a copy of the data or even a compressed copy can be impractical. We calculate likely spectrum signatures, or atoms, and calculate abundances which can allow classification when atoms are confirmed or calculated later. 

In Section \ref{sec:meth}, we describe the problem and solutions mathematically. 
In Section \ref{sec:exp}, we give experiments and results comparing matching pursuit, autoencoders, least squares, and least squares with radial basis function pruning versus noise, sparsity, and compression. 
Finally, in Section \ref{sec:concl}, we conclude.

\section{METHODOLOGY} \label{sec:meth}

\quad Using physical models, we have that electromagnetic waves, or photons, accumulate. For example, two emissions, $a_1$ and $a_2$ will be measured as $a_1 + a_2$. More generally, for many emissions of varying intensity, measurement $y = \sum_i c_i a_i$. We call the set of distinct emissions $\{a_i\}_i$ the atoms which make up the dictionary (please note that this definition of atom does not correspond to the physics/chemistry definition of atom). To accommodate noise or missing atoms, given measurement $y$ and the atoms, we can minimize $\|y - \sum_i c_i a_i\|$ over coefficients $\{c_i\}_i$. The set of atoms might be very large while we expect few $c_i$ to be nonzero. This sparsity constraint can be incorporated by optimizing $\min_{\vec c} \|y - A \vec c\| + \lambda \| \vec c \|_1$ where matrix $A = [a_1,a_2,...]$ and $\vec c = [c_1,c_2,...]^T$. However, optimizing this is computationally expensive so pruning is often used instead. We call $A$ the dictionary. To prune the dictionary, atoms, that is columns, are removed. Then with the pruned dictionary, $\|y - A \vec c\|$ is minimized and the resulting $\vec c$ is at least as sparse (number of nonzero elements) as the pruned dictionary. The matching pursuit\cite{Mallat} (MP) method also computes $\vec c$ with a given sparsity to 
\begin{equation} \label{eq:vec_c}
 \texttt{minimize}_{\vec c} \ \|y - A \vec c\|, 
\end{equation}
but is only optimal, or even convergent, when the atoms are orthogonal, that is $\langle a_i, a_j \rangle = a_i^T a_j = 0$ for all $i \ne j$. 

The standard pruning method removes the atoms least correlated with the measurement where correlation is the inner product. Let pruned dictionary $\widehat A = [a_{i_1}, a_{i_2},...]$ where $|\langle a_{i_j}, y \rangle| > \delta$ for all $j$, see Algorithm \ref{alg:pnnls} Line 15. We assume the atoms are normalized, meaning $\langle a_i, a_i \rangle = 1$ for all $i$, otherwise rescale them. This normalization to the sphere does not affect accuracy since the coefficient can account for the scale change. We are assuming that no atoms are $\vec 0$ or approximately $\vec 0$. This is also called a subspace projection. 

We developed Hilbert space pruning as follows. Let $\phi$ map atoms in $\mathcal{M}$ to a reproducing kernel Hilbert space called $\mathcal{H}$; $\phi : \mathcal{M} \rightarrow \mathcal{H}$. Hilbert space pruning removes the atoms with least Hilbert correlation with the measurement. The pruned dictionary $\widehat A = [a_{i_1}, a_{i_2},...]$ where $|\langle \phi( a_{i_j}), \phi( y ) \rangle_{\mathcal{H}}| > \delta$ for all $j$. Now when $\phi$ is the identity and $\mathcal{M} = \mathcal{H}$, Hilbert pruning is standard pruning. However, the reproducing kernel and ``the kernel trick'' gives us that $k(x,y) = \langle \phi(x), \phi(y) \rangle_H$. We choose $k$ to be the Gaussian radial basis function. Then $\phi$ and $\mathcal{H}$ exist without us needing to choose or calculate them and 
$$k(x,y) = \exp(-\gamma\|x-y\|_2^2).$$ 
Then $\widehat A = [a_{i_1}, a_{i_2},...]$ where $\exp(-\gamma\|a_{i_j}-y\|_2^2) > \delta$ for all $j$, see Algorithm \ref{alg:pnnls} Line 18. We will assume $\{a_i\}_i$ and $y$ are normalized to 1. We give the pseudocode in Algorithm \ref{alg:pnnls}. 

\begin{algorithm}[H]
\caption{Projected Nonnegative Least Squares (PNNLS) Standard/RBF Pruning}\label{alg:pnnls}
\textbf{Input:} 

\quad Y : p $\times$ M $\times$ M - data array

\quad p : number of frequencies

\quad M : number of pixels along each axis

\quad A : p $\times$ N - dictionary array

\quad N : number of atoms

\quad k : sparsity integer

\textbf{Output:} 

\quad $\Gamma$ : N $\times$ M $\times$ M - abundance data array 

\textbf{Begin:} \\

$\Gamma = 0$

\For{$i,j : 1 \le i,j \le M$}{
    $y = Y[:,i,j]$
    
    \If{standard pruning}{
        $ T = abs(A^T y) $
    }
    \If{RBF pruning}{
        \For{$k : 1 \le k \le N$}{
            $ T[k] = \exp(-\gamma\| A[:,k] - y\|_2^2 ) $ 
        }
    }
    
    $ I : I \subset [1,N], \ |I|=k, \ T_I \ge T_{[1,N]\backslash I} $ \quad \# Highest subset
    
    $c = \arg\min_c \| y - A_I c \|_2^2$
    
    $\Gamma[I,i,j] = c$
}
\end{algorithm}

Sparse abundance vectors, $\vec c$, compress measurements, $y$. Unfortunately, when measurements have high resolution or vector $y$ has many indices, the space of measurements is exponentially larger. This makes the representation space, or the span of the atoms, less likely to be close to measurement $y$. We call this difference, $\min_{\vec c}\|y - A \vec c \|$, the remaining variance after fixing $A$. We find that the remaining variance of a dataset or image is efficiently compressed with the singular vectors of the remaining variance data matrix. By adding these singular vectors to the dictionary of atoms, the representation, or compression, error is greatly reduced. We call the vectors added to the dictionary, compression vectors, see Algorithm \ref{alg:compression}. 

\begin{algorithm}[h]
\caption{Abundance Calculation with Compression Vectors}\label{alg:compression}
\textbf{Input:} 

\quad Y : p $\times$ M $\times$ M - data array

\quad p : number of frequencies

\quad M : number of pixels each axis

\quad A : p $\times$ N - dictionary array

\quad N : number of atoms

\quad k : sparsity integer

\quad c : number of compression vectors

\textbf{Output:} 

\quad $U$ : $p \times c$ - compression vectors

\quad $\Gamma$ : (N+c) $\times$ M $\times$ M - abundance data array 

\textbf{Begin:} \\

$\Gamma = $Algorithm1(Y,p,M,A,N,k)

\For{$i,j : 1 \le i,j \le M $}{
    $Y_{variance}[:,i,j] = Y[:,i,j] - A \ \Gamma[:,i,j]$
}

$Y_{variance\_reshape} =$ Reshape($Y_{variance}$,p,$M^2$)

$[U,D,V] = SVD(Y_{variance\_reshape})$

$U = U[:,1:c]$ \quad \# Highest variance vectors

$\Gamma_{compression} = $Algorithm1($Y_{variance},M,U,N,c$)

$\Gamma = $concat($\Gamma, \Gamma_{compression}$)
\end{algorithm}

\section{EXPERIMENTS}\label{sec:exp}
We utilize two datasets. One is the Urban\cite{urban} dataset and the other is a spectral library from USGS\cite{usgs}.

\subsection{USGS Spectral Library}

\quad We synthesize hyperspectral images for experiments with real spectra. The USGS dataset has thousands of spectrums and we choose the 448 
ASD spectrums under \texttt{ChapterM\_Minerals} in \texttt{ASCIIdata\_splib07a}, as atoms for our experiments. 
We synthesize images pixel by pixel. For each pixel, we take a random positive vector $\vec\beta$ and treat this as the abundance ground truth. Then we mix the atoms with it to create measurement $y=A\vec\beta$. The method then calculates $\vec c$ from Equation \ref{eq:vec_c} and the error is 
\begin{equation}\label{eq:rec_error}
E_{A,\beta,y} = \|\vec c-\vec\beta\|.
\end{equation}



\subsubsection{Signal-to-Noise}

\begin{figure}[H]
\centering
  \includegraphics[width=0.85\textwidth]{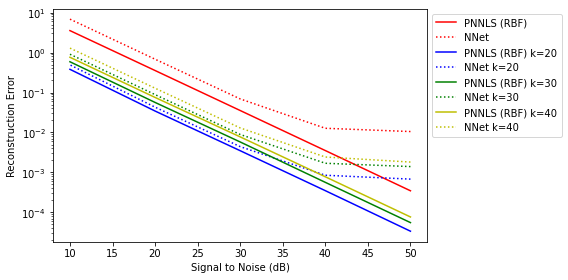}
\caption{Reconstruction L1 error, Equation \ref{eq:rec_error}, over hyperspectral image, of projected non-negative least squares (PNNLS RBF) versus autoencoder (NNet) plotted versus measurement noise level. k is sparsity or number of nonzero variables. Each curve is a mean of 10 curves from 10 I.I.D. synthetic images from USGS spectral library \cite{usgs}. Notice that both axes are log scaled. 
}
\label{fig:noise_nnet}
\end{figure}

We compared a neural network autoencoder (Nnet), as in Wang et. al.{\cite{Autoencoder}}, with PNNLS RBF with varying signal to noise ratios where $y = A\vec\beta + \eta$ and $\eta$ is Gaussian noise. The Nnet has fully connected layers of size 10k/ReLu, 5k/ReLu, 2k/ReLu and then embeds into the latent space of atom coefficients. Then the next layer is set to be the corresponding dictionary atoms which are combined into the output layer (same dimension as input layer). 
We test sparseness $k$ at 20, 30, and 40. The measurement sparseness is lower and is composed of a random 17 atoms of the dictionary containing 448 atoms from USGS. All dictionary atoms are orthogonalized with Gram-Schmidt so that the solution is well-posed. In Figure \ref{fig:noise_nnet}, we see that as SNR increases, PNNLS has error converging to 0. However, NNet has increasing accuracy up to 40 dB but then fails to converge. For $k \leq 17$, PNNLS and NNet fail to converge and have high error. This is expected because $<17$ atoms are not enough to represent the 17 atoms in the measurement. Our experiment consists of 10 pixels where each is a unique random linear combination. 

\subsubsection{Sparsity}

\begin{figure}[H]
\centering
  \includegraphics[width=0.7\textwidth]{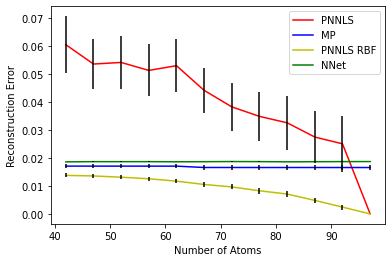}
\caption{Reconstruction L1 error, Equation \ref{eq:rec_error}, averaged over atoms, plotted versus abundance vector sparsity, which is the number of atoms with nonzero coefficients. Black bars are 1/5 standard deviation of 30 curves of 30 I.I.D. samples. Spectrum data from USGS spectral library \cite{usgs}.
}
\label{fig:atoms}
\end{figure}

Next we compare the reconstruction L1 error versus imposed sparsity. We test upon 97 atoms from USGS dataset. We compare projected non-negative least squares (PNNL) with standard pruning, PNNL with Hilbert space pruning (RBF), and deep learning autoencoders (NNet) as in Wang et al{\cite{Autoencoder}}. The noise model here multiplies each atom with a binomial random variable that is 1 with 90\% chance and -1 with 10\% chance. As the number of nonzero coefficients increases, the error decreases. Hilbert space pruning reduces the error by up to 40\% in this experiment compared to standard pruning, see Figure \ref{fig:atoms}. Hilbert space pruning and standard pruning have error converging to zero at 100 atoms which is the number of atoms actually composing the measurement. The deep learning autoencoder has higher error and does not converge because gradient descent gets trapped in local minimums with non-convex optimization. We also plot the standard deviation information where we see that MP, PNNLS RBF, and NNet are relatively stable but PNNLS is unstable. 

\subsection{Urban Dataset}

\quad The Urban dataset from the HYDICE sensor consists of a hyperspectral image with 94,249 pixels and 162 spectral bands (after reduction) and 6 endmembers which have been identified as corresponding to the pervailing materials in the scene; we will use these endmembers as our dictionary atoms.
The bands are from 0.4 to 2.5 microns with resolution of 10 nm. Bands 1-4, 76, 87, 101-111, 136-153, and 198-210 were removed due to dense water vapor and atmospheric effects. 

\begin{figure}[h]
\centering
  \includegraphics[width=0.5\textwidth]{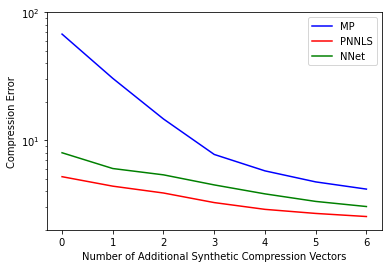}
\caption{Compression L1 error, averaged over pixels, of matching pursuit and projected non-negative least squares (PNNLS RBF) versus number of additional compression vectors used. Analysis of Urban dataset\cite{urban}.}
\label{fig:compression}
\end{figure}

Last but not least, we test the effect of using compression vectors with matching pursuit (MP), neural network autoencoder (NNet), and non-negative least squares (PNNLS). The compression error is $\|y - A \vec c\|_1/N$ where $y$ is the measurement, $A$ is the dictionary of atoms, $\vec c$ is the abundance vector, and $N$ is the dimensionality of $y$ (in this case 162). In Figure \ref{fig:compression}, we see that adding the compression vectors calculated with SVD (see Algorithm \ref{alg:compression}) captures the remaining variance of the data and decreases the compression error. However, PNNLS always has much less compression error and even with additional compression vectors, the error is consistently below matching pursuit and NNet.

\section{CONCLUSION} \label{sec:concl}
\quad We have shown that projected non-negative least squares with RBF (PNNLS RBF) is more noise tolerant than matching pursuit (MP) and the autoencoder (NNet) in experiments varying the SNR. Then we showed that a Hilbert transformation in pruning, ex: PNNLS RBF, reduces reconstruction error in correlated noise regimes. PNNLS RBF also achieved lower error than neural network autoencoder methods in our testing. Finally, we introduced compression vectors as a way to efficiently capture signals for later improved classification. We found that because PNNLS RBF has lower reconstruction error than MP and NNet, its compression error is lower. We found that even after adding compression vectors, PNNLS RBF still had lower compression errors. 

\acknowledgments 

\quad This work was conducted under the Laboratory Directed Research and Development Program at PNNL, a multi-program national laboratory operated by Battelle for the U.S. Department of Energy under contract DE-AC05-76RL01830.

\bibliography{report} 
\bibliographystyle{spiebib} 

\end{document}